\documentclass[letterpaper, 10 pt, conference]{ieeeconf}
\IEEEoverridecommandlockouts
\usepackage[vlined, ruled, boxed, linesnumbered]{algorithm2e}

\SetCommentSty{mycommfont}

\SetKwProg{Function}{}{}{}

\usepackage{graphicx}
\usepackage{graphics}
\usepackage{times}
\usepackage{amsmath}
\usepackage{amssymb}
\usepackage{float}
\usepackage{url}
\usepackage{subfigure}
\usepackage{caption}
\usepackage{multirow}
\usepackage{gensymb}
\usepackage{epstopdf}
\usepackage{wrapfig}
\usepackage{setspace} 
\usepackage[noend]{algpseudocode}
\usepackage{color}
\usepackage{cite}
\usepackage{booktabs}
\usepackage{stackengine}
\usepackage[font={small}]{caption}

\usepackage{tikz}
\usepackage{textcomp}
\usepackage{lipsum}

\title{\LARGE \bf
A Receding Horizon Multi-Objective Planner \\ for Autonomous Surface Vehicles in Urban Waterways
}

\author{Tixiao Shan, Wei Wang, Brendan Englot, Carlo Ratti, and Daniela Rus
\thanks{
T. Shan, W. Wang, and C. Ratti are with the Department of Urban Studies and Planning, Massachusetts Institute of Technology, USA, {\tt\scriptsize \{shant, wweiwang, ratti\}@mit.edu}. \newline
\indent B. Englot is with the Department of Mechanical Engineering, Stevens Institute of Technology, USA, {\tt\scriptsize benglot@stevens.edu}. \newline
\indent T. Shan, W. Wang and D. Rus are with the Computer Science \& Artificial Intelligence Laboratory, Massachusetts Institute of Technology, USA, {\tt\scriptsize \{shant, wweiwang, rus\}@mit.edu}.}%
}

\begin{document}

\maketitle
\thispagestyle{empty}
\pagestyle{empty}


\begin{abstract}
We propose a novel receding horizon planner for an autonomous surface vehicle (ASV) performing path planning in urban waterways. 
Feasible paths are found by repeatedly generating and searching a graph reflecting the obstacles observed in the sensor field-of-view. We also propose a novel method for multi-objective motion planning over the graph by leveraging the paradigm of \textit{lexicographic optimization} and applying it to graph search within our receding horizon planner. The competing resources of interest are penalized hierarchically during the search. Higher-ranked resources cause a robot to incur non-negative costs over the paths traveled, which are occasionally zero-valued. The framework is intended to capture problems in which a robot must manage resources such as risk of collision. This leaves freedom for tie-breaking with respect to lower-priority resources; at the bottom of the hierarchy is a strictly positive quantity consumed by the robot, such as distance traveled, energy expended or time elapsed. We conduct experiments in both simulated and real-world environments to validate the proposed planner and demonstrate its capability for enabling ASV navigation in complex environments.

\end{abstract}


\section{Introduction}
\label{sec::introduction}

Great efforts have been devoted to enhancing the capabilities of autonomous surface vehicles (ASVs) in the last few decades. Among them, the recently launched Roboat project seeks to explore the complex interactions between human society and ASVs \cite{johnsen2019roboat}. The Roboat project aims to provide water-based transportation to relieve the congestion of saturated road-based transportation in Amsterdam, the Netherlands. Such water-based transportation includes but is not limited to applications such as public transportation, waste collection, and package delivery. The deployment of Roboat seeks to contribute novel urban infrastructure that supports the development of a modern city.

Deploying an autonomous boat in the busy canals of a major city involves designing systems for perception, localization, and path planning. Among them, path planning in particular plays a crucial role in enabling safe navigation of urban canals, as its outcome directly influences the interactions between the vehicle and its surrounding environment. Various challenges can be foreseen when deploying such a system. Cruising in urban waterways is subject to rules that are similar to driving on roadways. Autonomous vehicles may only be licensed to cruise in certain regions of the canal, while following a pre-defined route from the  relevant authorities. In addition, the behavior of the vehicles shouldn't disrupt the course of human-controlled boats. 

\begin{figure}[t!]
	\centering
	\subfigure[Tourism transportation]{\includegraphics[width=.4\columnwidth]{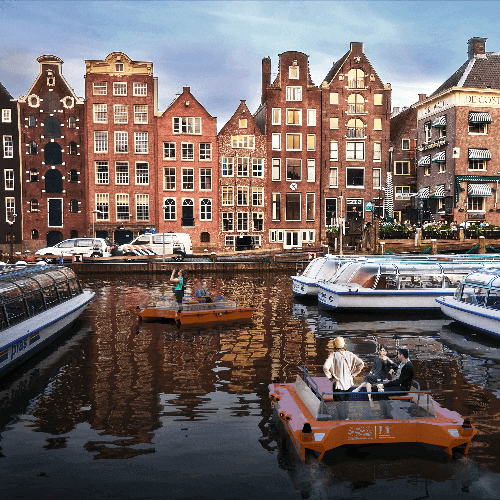}}
	\subfigure[Autonomous taxi]{\includegraphics[width=.4\columnwidth]{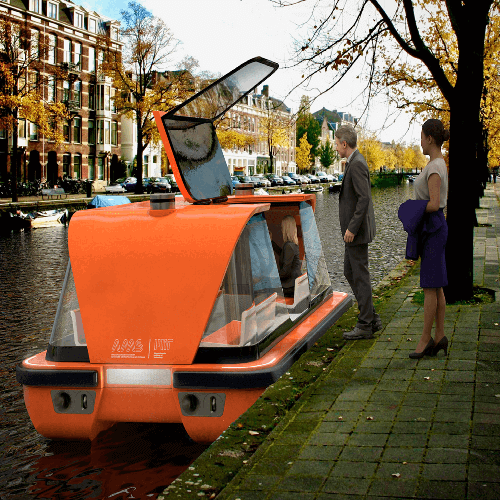}}
	\subfigure[Waste collection]{\includegraphics[width=.4\columnwidth]{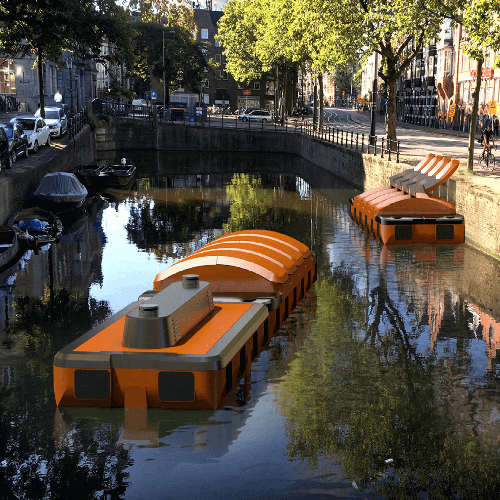}}
	\subfigure[Package delivery]{\includegraphics[width=.4\columnwidth]{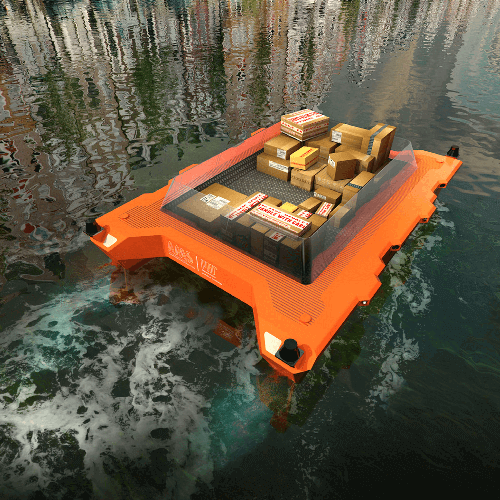}}
	\caption{Representative applications of Roboat: (a) tourism transportation, (b) public transportation, (c) waste collection, (d) package delivery.}
	\label{fig::roboat-demo}
	\vspace{-5mm}
\end{figure}

With these challenges in mind, this paper focuses on developing a path planning algorithm that enables ASV navigation in complex urban waterways. The proposed planner is designed according to a receding horizon scheme, repeatedly generating and searching a graph that reflects the obstacles observed in the sensor field-of-view. We model the various challenges to be solved during navigation as costs to be minimized. Then the planning problem can be treated as a multi-objective optimization problem. We propose a lexicographic search algorithm to solve this multi-objective planning problem quickly without parameter-tuning, by ranking the objectives hierarchically. The main contributions of our work can be summarized as follows:
\begin{itemize}
	\item A novel receding horizon planner that is suitable for autonomous navigation of ASVs in urban waterways.
	\item An efficient, multi-objective search algorithm that enables real-time performance without iterative adjustment of constraints by hierarchically ranking the objectives.
	\item The proposed framework is validated with tests in both simulated and real-world environments.
\end{itemize}

\section{Related Work}
\label{sec::related-work}

The most relevant body of prior work is in multi-objective motion planning. 
In pursuit of solutions that can be produced quickly, preferably in real-time, and applied to problems of high dimension, sampling-based motion planning algorithms such as the probabilistic roadmap (PRM) \cite{kavraki1994}, the rapidly-exploring random tree (RRT) \cite{lavalle2001}, and their optimal variants PRM*, RRT*, and rapidly-exploring random graphs (RRG) \cite{RRT*} have been adapted to solve a variety of multi-objective motion planning problems. Such approaches have typically considered the tradeoff between a resource such as time, energy, or distance traveled and a robot's information gathered \cite{hollinger2014}, localization uncertainty \cite{bopardikar2015}, \cite{shan2017belief}, collision probability \cite{roy2011}, clearance from obstacles \cite{kim2003}, adherence to rules \cite{reyes2013},  and exposure to threats \cite{clawson2015}. 

We consider problems in which two or more competing resources are penalized \textit{hierarchically}. The higher-priority resources assume non-negative costs over robot paths, and are frequently zero-valued. This is intended to capture problems in which robots must manage resources such as collision risk, access to valuable measurements or following certain rules, which are present in some regions of the environment, and absent in others. For example, \cite{shan2015sampling} proposed a sampled-based planning algorithm for minimum risk planning. Risk is only penalized in the regions of the environment where collision is possible. This leaves freedom for tie-breaking with respect to a secondary resource, such as distance traveled. A min-max uncertainty planning algorithm is proposed in \cite{englot2016sampling} for planning under uncertainty. When the primary cost, localization uncertainty, is not increasing, a secondary and a tertiary cost are introduced to break ties. These planning problems fit nicely into the framework of \textit{lexicographic optimization}. 

The lexicographic method \cite{stadler1988} is the technique of solving a multi-objective optimization problem by arranging each of several cost functions in a hierarchy reflecting their relative importance. The objectives are minimized in sequence, and the evolving solution may improve with respect to every subsequent cost function if it does not worsen in value with respect to any of the former cost functions. Use of this methodology has been prevalent in the civil engineering domain, in which numerous regulatory and economic criteria often compete with the other objectives of an engineering design problem. Variants of the lexicographic method have been used in land use planning \cite{veith2003}, for vehicle detection in transportation problems \cite{sun1999}, and in the solution of complex multi-objective problems, two criteria at a time \cite{engau2007}. 

Among the benefits of such an approach is the potential for the fast, immediate return of a feasible solution that offers globally optimal management of the primary resource, in addition to locally optimal management of secondary resources in areas where higher-ranked resources are zero-valued. Due to the fact that the spatial regions in which resources are penalized can often be intuitively derived from a robot's workspace, using facts such as whether the robot is in an allowed operating region, or whether a robot is within range of communication or sensing resources, such an approach offers an intuitive means for managing the relative importance of competing cost functions, in which the user needs only to select the order in which the resources are penalized. This stands in contrast to methods that require tuning of additive weights on the competing cost functions \cite{reyes2013}, and robot motion planning methods that manage the relative influence of competing cost criteria using constraints, \cite{hollinger2014}, \cite{roy2011}. Avoiding any potential struggles to recover feasible solutions under such constraints, the lexicographic motion planning problem is \textit{unconstrained} with respect to the resources of interest. 

\section{Problem Definition}
\label{sec::problem-definition}

\subsection{Path Planning}
\label{sec::problem-planning}
Let $\mathcal{C}$ be a robot's configuration space. $x \in \mathcal{C}$ represents the robot's configuration. $\mathcal{C}_{obst} \subset \mathcal{C}$ denotes the set of configurations that are in collision with the obstacles, which are perceived from the sensor data $\mathcal{S}$. $\mathcal{C}_{free} = cl(\mathcal{C} \backslash \mathcal{C}_{obst})$, in which $cl()$ represents the closure of an open set, denotes the space that is free of collision in $\mathcal{C}$. We assume that given a current configuration $x_{c} \in \mathcal{C}_{free}$ and a global reference path $\mathcal{G}$, the robot must travel in $adj(\mathcal{G})$,  which represents the neighboring regions of $\mathcal{C}$ adjacent to $\mathcal{G}$, and reach a goal state $x_{g}$, which is located at the end of $\mathcal{G}$. 

Let a \textit{path} be a continuous function $\sigma : [ 0,1 ]  \rightarrow \mathcal{C}$ of finite length. Let $\Sigma$ be the set of all paths $\sigma$ in a given configuration space. A path $\sigma$ is collision-free and feasible if $\sigma \in \mathcal{C}_{free}$, $\sigma(0) = x_{c}$ and $\sigma(1) = x_{g}$. A feasible path $\sigma$ is composed of two segments, $\sigma = \sigma_{G} \cup \sigma_{\mathcal{G}}$. $\sigma_{G}$, which exists in $adj(\mathcal{G})$, is obtained by searching a directed graph $G(V,E)$, with node set $V$ and edge set $E$. $\sigma_{\mathcal{G}} \in \mathcal{G}$ is directly obtained from $\mathcal{G}$. $\sigma_{G}$ and $\sigma_{\mathcal{G}}$ can be concatenated as $\sigma_{G}(1)=\sigma_{\mathcal{G}}(0)$. An edge $e_{ij} \in E$ is a path $\sigma_{i,j}$ for which $\sigma_{i,j}(0) = x_i \in V$ and $\sigma_{i,j}(1) = x_j \in V$. Two edges $e_{ij}$ and $e_{jk}$ are said to be linked if both $e_{ij}$ and $e_{jk}$ exist. A path $\sigma_{p,q} \in G$ is a collection of linked edges such that $\sigma_{p,q} = \{e_{p \; i_1},e_{i_1 i_2}, ..., e_{i_{n-1} i_n}, e_{i_n q}\}$.
The problem of finding a feasible path may be specified using the tuple $({C}_{free},x_{c},\mathcal{G})$.

\subsection{Lexicographic Optimization}
\label{sec::problem-lexico}
We define cost functions $c_k(\sigma)$, where $c_k: \Sigma \rightarrow \mathbb{R}_{0}^{+}$ maps a path $\sigma$ to a $k$th non-negative cost,  $k \in \{1,2,...,K\}$, and $K$ is the total number of costs in a multi-objective planning problem. These $K$ cost functions are applied to the problem of lexicographic optimization \cite{marler2004}, which may be formulated over collision-free paths as
\begin{align}
\label{eq::lexico-formula}
&\sigma^* = \underset{\sigma_{k}\in \mathcal{C}_{free}}{\text{min}}{c_{k}(\sigma) }\;\;\;\; \\
&\text{subject to}: \; c_{j}(\sigma)\leq c_{j}(\sigma_{j}^{*})\;\; \nonumber \\
&\text{where}: j=1,2,...,k-1, k>1;\nonumber \\
&\;\;\;\;\;\;\;\;\;\;\;\; k=1,2,...,K. \nonumber
\end{align}
The formulation of the lexicographic method is adapted here (we refer the reader to the description from \cite{marler2004}, Section 3.3) to show cost functions that take collision-free paths as input. We also assume specifically that $c_K: \Sigma \rightarrow \mathbb{R}^{+}$, implying that $c_K$ increases monotonically over a path, as with costs such as distance traveled or time elapsed. Accordingly, ties rarely occur in the bottom level of the hierarchy, with their incidence depending on the algorithm adopted for path generation. In one iteration of the procedure of Equation (\ref{eq::lexico-formula}), a new solution $\sigma^*$ will be returned if it does not increase in cost with respect to any of the prior cost functions $j < k$ previously examined. Necessary conditions for optimal solutions of Equation (\ref{eq::lexico-formula}) were first established by Rentmeesters \cite{rentmeesters1996}. Relaxed versions of this formulation have also been proposed, in which $c_{j}(\sigma_{k}) > c_{j}(\sigma_{j}^{*})$ is permitted, provided that $c_{j}(\sigma_{k})$ is no more than a small percentage greater in value than $c_{j}(\sigma_{j}^{*})$. This approach, termed the \textit{hierarchical method} \cite{osyczka1984}, has also been applied to multi-criteria problems in optimal control \cite{waltz1967}.

\subsection{Cost Function}
\label{sec::problem-cost}
We introduce three types of costs, which are inspired by our application of Roboat in urban waterways, to demonstrate the usage of lexicographic optimization in our planning problem. The three costs, which are ranked hierarchically, are risk cost, heading cost, and distance cost. 

The risk accumulated along a path $\sigma$ is derived using:
\begin{align}
\label{eq::cost-risk}
c_{1}(\sigma) & := \int_{\sigma(0)}^{\sigma(1)} Risk(\sigma(s)) \;ds \\
Risk(x) & := \left\{ 
\begin{array}{lr}
\mathcal{R}(x) &\text{if} \; \mathcal{R}(x) > Th_{risk} \\
\; \; \; 0 &\text{otherwise} \; \; \; \; \; \; \; \; \; \; \; \;
\end{array}
\right.,
\end{align}
where the function $Risk()$ evaluates the risk at an individual robot state. Let us assume that 
$\mathcal{R}(x)$ is defined as the inverse of the distance between $x$ and the nearest obstacle to $x$. The $Risk()$ function is activated if $\mathcal{R}(x)$ is larger than a risk threshold $Th_{risk}$. For example, when we let $Th_{risk}=2$, $Risk()$ gives non-zero values when the robot is within 0.5 m of any obstacles. When we let $Th_{risk}=\infty$, $Risk()$ returns zero everywhere in $\mathcal{C}_{free}$. The logic behind employing this cost function is that we wish to place a \textit{comfort zone} between the robot and other surrounding objects, especially human-driven boats. The ASV should try to avoid this zone to minimize its influence on other vessels. Minimizing the risk cost results in minimizing the travel distance of the ASV in these zones. Another approach to create such a zone is to naively inflate the obstacle regions. However, a naive inflation of obstacles may block the entire waterway if they are close, even though there is a feasible path passing through them. An example of the proposed comfort zones, which are colored gray, is shown in Figure \ref{fig::pipeline}(a).

In addition, we define heading cost as the secondary cost, which penalizes the heading difference between the robot and the global reference path $\mathcal{G}$:
\begin{align}
\label{eq::cost-heading}
c_{2}(\sigma) & := \int_{\sigma(0)}^{\sigma(1)} Heading(\sigma(s)) \;ds \\
Heading(x) & := \left\{ 
\begin{array}{lr}
\mathcal{H}(x) &\text{if} \; \mathcal{H}(x) > Th_{head} \\
\; \; \; 0 &\text{otherwise} \; \; \; \; \; \; \; \; \; \; \;\;\;
\end{array}
\right.,
\end{align}
where the function $\mathcal{H}(x)$ gives the heading difference between $x$ and the heading of the path segment that is the closest to $x$ on $\mathcal{G}$. Due to wind and wave interference, aligning the heading of the robot with $\mathcal{G}$ perfectly is practically impossible. To avoid exhaustive control effort, we define a heading difference threshold $Th_{head}$. $Heading(x)$ returns a non-zero value when the error $\mathcal{H}(x)$ is larger than $Th_{head}$. Incorporating this cost ensures the generated path is relatively smooth while adhering to the heading of $\mathcal{G}$.

At last, we define travel distance as the tertiary cost, which is strictly positive. This ensures that ties do not occur here as frequently as they do for the primary or secondary cost. The distance cost is defined as follows:
\begin{align}
\label{eq::cost-dist}
c_{3}(\sigma) & := \int_{\sigma(0)}^{\sigma(1)} Distance(\sigma(s)) \;ds  \;.
\end{align}

The definition of these three costs is intended to support our deployment of Roboat in urban waterways. The risk cost minimizes the interference of Roboat with other objects to ensure a safe and comfortable ride for passengers on Roboat and other vehicles. The heading cost helps yield a smooth ride for the passengers while minimizing the control effort. The strictly positive distance cost ensures that ties rarely occur in the bottom level of the hierarchy, and minimizes the travel distance if possible.
We note that a user can incorporate other types of cost functions or rules into the cost hierarchy based on their importance in the application of interest.

\section{Receding Horizon Planner}
\label{sec::planner}
\begin{figure}[t!]
	\centering
	\subfigure[Reference path $\mathcal{G}$]{\includegraphics[width=.75\columnwidth]{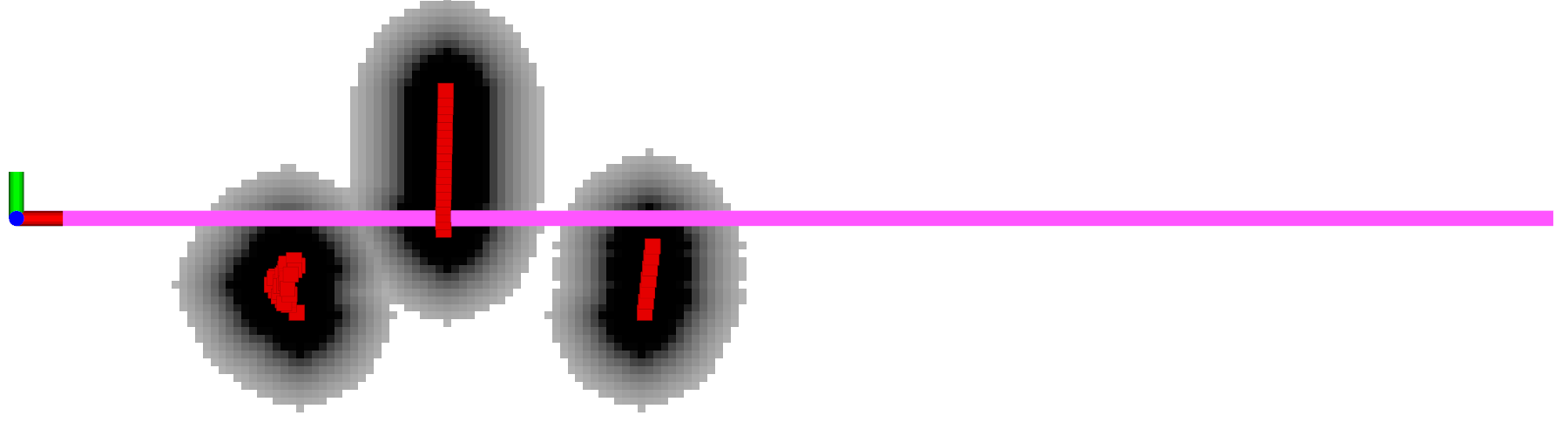}}
	\subfigure[Generated nodes $V$]{\includegraphics[width=.75\columnwidth]{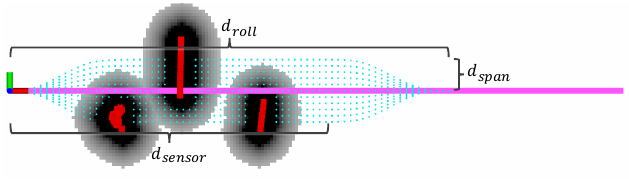}}
	\subfigure[Generated graph $G$ and feasible path $\sigma$]{\includegraphics[width=.75\columnwidth]{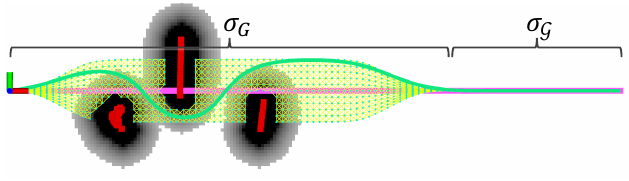}}
	\caption{Example of the proposed receding horizon planner. In (a), The robot is located on the left. The reference path $\mathcal{G}$ and obstacles are colored pink and red respectively. The black color around the obstacles indicates inflated obstacles taking into account the robot's size. The gray color indicates the risk region defined in Section \ref{sec::problem-cost}. In (b), the graph $G$ is generated due to obstacles intersecting $\mathcal{G}$. In (c), a feasible path $\sigma = \sigma_{G} \cup \sigma_{\mathcal{G}}$ is found, where $\sigma_{G}$ is returned by searching the graph $G$. $\sigma_{\mathcal{G}}$ is directly obtained from $\mathcal{G}$.}
	\label{fig::pipeline}
	\vspace{-3mm}
\end{figure}

To extend the application of the proposed planner to platforms with limited computational resources, we design our planner in a receding horizon manner. Given the global path $\mathcal{G}$ as the reference path, we only search for a new path in $adj(\mathcal{G})$ under necessary conditions, such as obstacles lying on $\mathcal{G}$ or the path to be executed. We also assume a prior map of the environment is not available for our planner, apart from the basic topological information required to formulate the reference path $\mathcal{G}$. This is because the environment of an urban waterway changes constantly due to human activities, such as canal maintenance and moving boats. A detailed understanding of the robot's environment is achieved by using the real-time perceptual sensor data $S$. With no requirement for a prior map, the proposed planner may be readily deployed. 

The pipeline of the proposed receding horizon planner is introduced in Algorithm \ref{alg::pseudo-algorithm}. The planner takes the robot's current state $x_{c}$, a global reference path $\mathcal{G}$, and the perceptual sensor data $S$ as inputs. When the global path $\mathcal{G}$ is not fully executed by the robot, we check the feasibility of path $\sigma$ using the perceived sensor data $S$. Note that we let $\sigma = \mathcal{G}$ when $\mathcal{G}$ is received at the beginning of the planning process. An illustrative example of the proposed planner is shown in Figure \ref{fig::pipeline}, with its planning horizon dictated by key parameters $d_{span}$, $d_{roll}$, and $d_{sensor}$, described below.

\begin{algorithm}
\caption{Receding Horizon Planner}
\label{alg::pseudo-algorithm}
\KwIn{robot state $x_{c}$, Global path $\mathcal{G}$, perceptual sensor data $\mathcal{S}$}
\While{$\mathcal{G}$ is not executed}
{
	Update obstacles from sensor data \\
	\If{new path $\sigma$ is needed}
	{
		Generate $G = (V,E)$ in $adj(\mathcal{G})$ from $x_{c}$\\
		Perform lexicographic search on $G$ \\
		\uIf{search is successful} {
		    \KwOut{path $\sigma = \sigma_{G} \cup \sigma_{\mathcal{G}}$}
		}
		\Else {
		    \KwOut{current robot state $x_{c}$}
		}
	}
}
\end{algorithm}

If obstacles are detected on $\sigma$, we then generate candidate robot states $V$ using $\mathcal{G}$ as the reference. We adapt the method introduced in \cite{darweesh2017open} and propose a \textit{roll-out} and \textit{roll-in} generation approach for sampling $V$. The states from the roll-out generation need to satisfy the robot's dynamic constraints while diverging from $x_{c}$ and spanning $adj(\mathcal{G})$. During the roll-in generation stage, the sampled states converge to $\mathcal{G}$ while remaining kinodynamically feasible. $adj(\mathcal{G})$ is defined by $d_{span}$ and $d_{roll}$. $d_{span}$ is the maximum distance between the sampled states and $\mathcal{G}$.  The total distance between the roll-out and roll-in sections is denoted as $d_{roll}$. In practice, $d_{roll}$ is set to be larger than the robot's sensor range $d_{sensor}$ in case the roll-in section converges on obstacles. Note that $V$ is sampled using a fixed density for the purpose of clear visualization. Figure \ref{fig::pipeline}(b) shows the generated $V$ with a fixed density of 0.1m. We then connect the states in $V$ and obtain a graph $G=(V,E)$. We only connect a state to its eight immediate neighboring states here for the purpose of visualization and show the resulting graph in Figure \ref{fig::pipeline}(c). 

\begin{algorithm}
\caption{Lexicographic Search}
\label{alg::lexicographic-search}
\KwIn{$G=(V,E), x_{init}, x_{goal}$}
$X_{queue} \leftarrow \{V\};$

\For{$k=1\:to\:K$}{
	$SetCost_{k}(V,\infty)$\;
	$x_{init}.c_{k} \leftarrow 0$\;
}
$x_{init}.parent \leftarrow \{\}$\;

\While{$|X_{queue}| > 0$}{
	$X_{min}\leftarrow X_{queue}$\;
	
   \For{$k=1\:to\:K$}{
		$X_{min}\leftarrow FindMinCost_{k}(X_{min})$\;
		\If{$|X_{min}| = 1$}{
			$x_{i} \leftarrow X_{min}$\;
			$break$\;
		}
	}
	
	$X_{queue} \leftarrow Pop(X_{queue}, x_{i})$\;
	
	\For{$\{x_{j} \; | \;  e_{ij}\in E \; and \; x_{j}\notin \sigma_{init,i})\}$}{
		\For{$k=1\:to\:K$}{
			\uIf{$c_{k}(x_{i},x_{j}) < x_{j}.c_{k}$}{
				\For{$n=k\:to\:K$}{
            		$x_{j}.c_{n} \leftarrow c_{n}(x_{i},x_{j})$\;
            	}
            	$x_{j}.parent \leftarrow x_{i}$\;
				$break$\;
			}
			\uElseIf{$c_{k}(x_{i},x_{j})=x_{j}.c_{k}$}{
				$continue$\;
			}
			\Else{	$break;$}
		}
	}
}
\end{algorithm}

We adapt Dijkstra's algorithm \cite{dijkstra1959} to perform a lexicographic graph search on $G$, which is detailed in Algorithm \ref{alg::lexicographic-search}. Provided with a graph $G(V,E)$ and $x_{init}$ as inputs, a queue $X_{queue}$ is populated with the nodes of the graph (Line 1), and the algorithm initializes $K$ cost-to-come costs for each node  (Lines 2-4). Each of these costs describes the $k$th priority cost-to-come for a respective node, along the best path identified so far  per the ranking of cost functions in Equation (\ref{eq::lexico-formula}). In real-time applications of the search, $x_{init}$ is designated to be the closest configuration in the graph to the robot's current state $x_{c}$, and $x_{goal}$ is the state in $\mathcal{G}$ that the graph $G$ converges to as roll-in occurs.

In each iteration of the algorithm's while loop, the $FindMinCost_k()$ operation returns the set of configurations that share the minimum $k$th priority cost-to-come from among the nodes provided as input (Line 9). If $X_{min}$ contains more than one configuration, lower-priority costs for the nodes in this set are examined until the set $X_{min}$ contains a single node, whose neighbors are examined in detail. The selected node is designated $x_i$ (Line 11). Node $x_{i}$ is then used, if possible, to reduce the costs-to-come associated with neighboring nodes $x_{j}$, if edge $e_{ij}$ exists. In Line 16, if $c_{k}(x_{i},x_{j})$, which represents the $k$th priority cost from $x_{init}$ to $x_{j}$ via $x_{i}$, is lower than the current cost, $x_{j}.c_{k}$, the costs-to-come of $x_{j}$ are updated by choosing $x_{i}$ as its new parent (Line 17-19). If, however, the $k$th priority cost from $x_{init}$ to $x_{j}$ via $x_{i}$ is tied with the current cost, $x_{j}.c_{k}$ (Line 21), then Algorithm \ref{alg::lexicographic-search} proceeds to the lower-priority cost $k+1$ and evaluates the potential $(k+1)$th priority cost-to-come improvements at $x_j$ by traveling via $x_i$. To reduce the likelihood of end-stage ties, the lowest-priority cost $K$ is assumed to be strictly positive over all paths in the configuration space.


Just as the problem formulation in Equation (\ref{eq::lexico-formula}) only allows improvements to a solution's lower-priority cost when it does not adversely impact a higher-priority cost, the proposed search method only allows improvements to be made in lower-priority costs when ties occur with respect to higher-priority costs. The single-source shortest paths solution produced by Algorithm \ref{alg::lexicographic-search} would take on the same primary cost whether or not these improvements are performed, but the occurrence of ties allows us to opportunistically address auxiliary cost functions in the style of lexicographic optimization.

\section{Algorithm Complexity}
\label{sec::algorithm-complexity}
The proposed lexicographic search, per the pseudo-code provided in Algorithm \ref{alg::lexicographic-search}, takes on worst-case complexity $O(K|V|^2)$. For clarity and illustrative purposes, we have used a naive $O(|V|^2)$ implementation of Dijkstra's algorithm, describing the lexicographic search using a basic queue that could be implemented using a linked list or similar. In the worst case, (1) finding the node(s) in the queue with the minimum cost (Line 9, costing $O(|V|^2)$ over the duration of the standard algorithm), and (2) expanding a node and inspecting its adjacent neighbors (Line 14, costing $O(|E|)$ over the duration of the standard algorithm) will each be repeated $K$ times, once for each cost function in the hierarchy, during every execution of the while loop. 

In the most time-efficient known implementation of Dijkstra's algorithm, which uses Fibonacci heaps \cite{fredman1987}, the complexity of the standard, single-objective algorithm is reduced from $O(|V|^2)$  to $O(|V|log|V| + |E|)$. Finding the minimum cost in the graph is trivial due to the maintenance of a priority queue, but deleting a node from the heap is a $O(log|V|)$ operation that must be repeated $|V|$ times over the duration of the algorithm. Expanding a node and inspecting its adjacent neighbors continues to cost $O(|E|)$ over the duration of the algorithm, since a worst case of $O(|E|)$ cost updates must be performed in the heap, each of which costs $O(1)$. To adapt this to a lexicographic search, the nodes in the heap must be prioritized per the lexicographic ordering of the graph nodes, so that the minimum cost reflects not only the minimum primary cost, but the optimum according to the formulation given in (1). Although only one node will be deleted from the heap in each iteration of the algorithm's while loop, each of the $O(log|V|)$ comparisons required will take $O(K)$ time, and so the cost of node deletion over the duration of the algorithm will increase to $O(K|V|log|V|)$.

The costs in the heap will also reflect the $K$ cost functions being considered. To maintain a lexicographic ordering among the nodes in the heap, all nodes undergoing cost changes during an iteration of the algorithm's while loop may have their costs individually adjusted as many as $K$ times. Akin to the steps performed in lines 15-20 of Algorithm \ref{alg::lexicographic-search}, this is the worst-case number of times a node's cost must be adjusted to establish the correct lexicographic ordering among a set of nodes with $K$ cost functions. Over the duration of the algorithm, this will result in a worst-case $O(K|E|)$ cost changes within the heap, each of which carries $O(1)$ complexity. As a result, the worst-case complexity of a lexicographic search using a Fibonacci heap will be improved to $O(K|V|log|V| + K|E|)$, from the original $O(K|V|^2)$. In the results to follow, we opt to implement and study the $O(K|V|^2)$ version of the algorithm in software, due to its ease of implementation and efficient memory consumption.

We also note briefly that an adaptation of Dijkstra's algorithm is selected in this application due to the fact that all graphs considered are characterized by non-negative, time-invariant edge weights. The consideration of negative edge weights would require an adaptation of the Bellman-Ford or Floyd-Warshall algorithm \cite{clrs}, and the consideration of time-varying weights, such as those which might depend on the action or measurement history of a robot, as frequently occurs in belief space planning, may require search algorithms of exponential complexity \cite{shan2017belief}.

\begin{figure}[t!]
	\centering
	\subfigure[]{\includegraphics[width=.75\columnwidth]{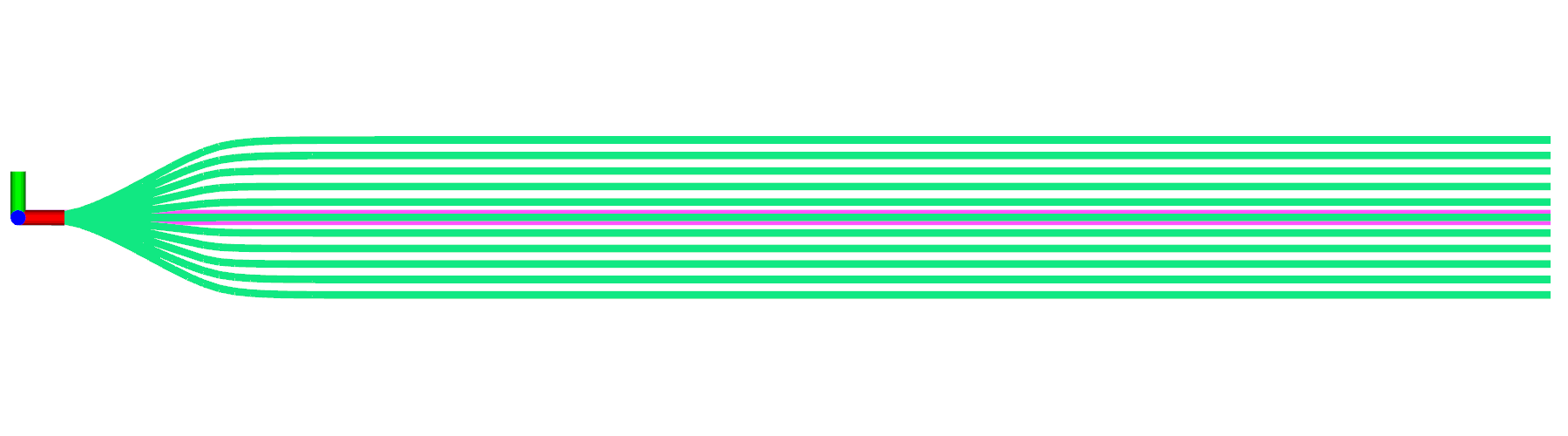}}
	\subfigure[]{\includegraphics[width=.75\columnwidth]{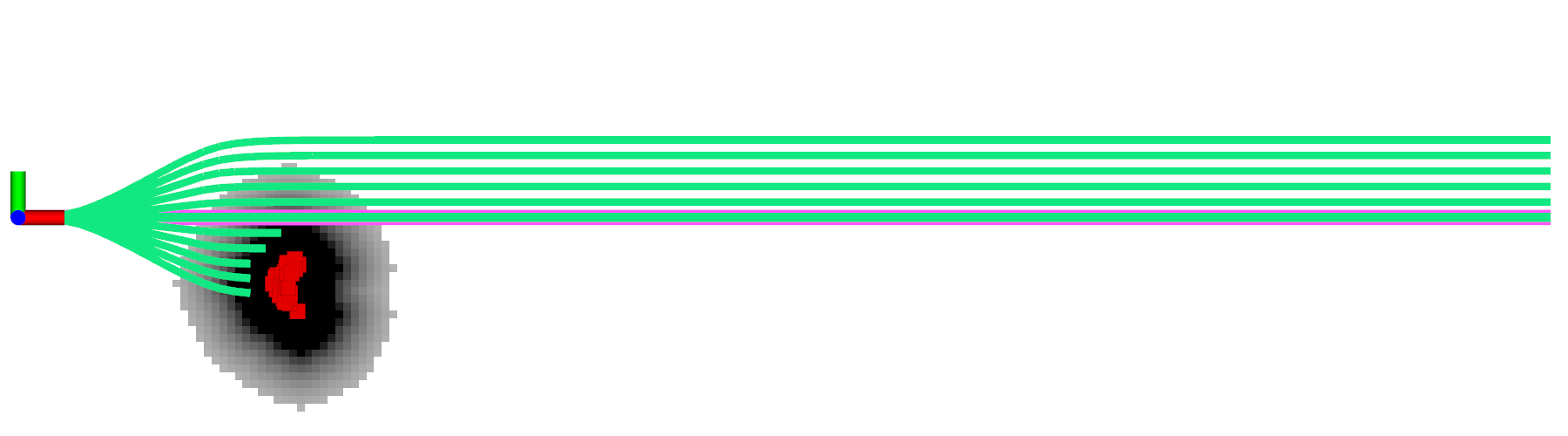}}
	\subfigure[]{\includegraphics[width=.75\columnwidth]{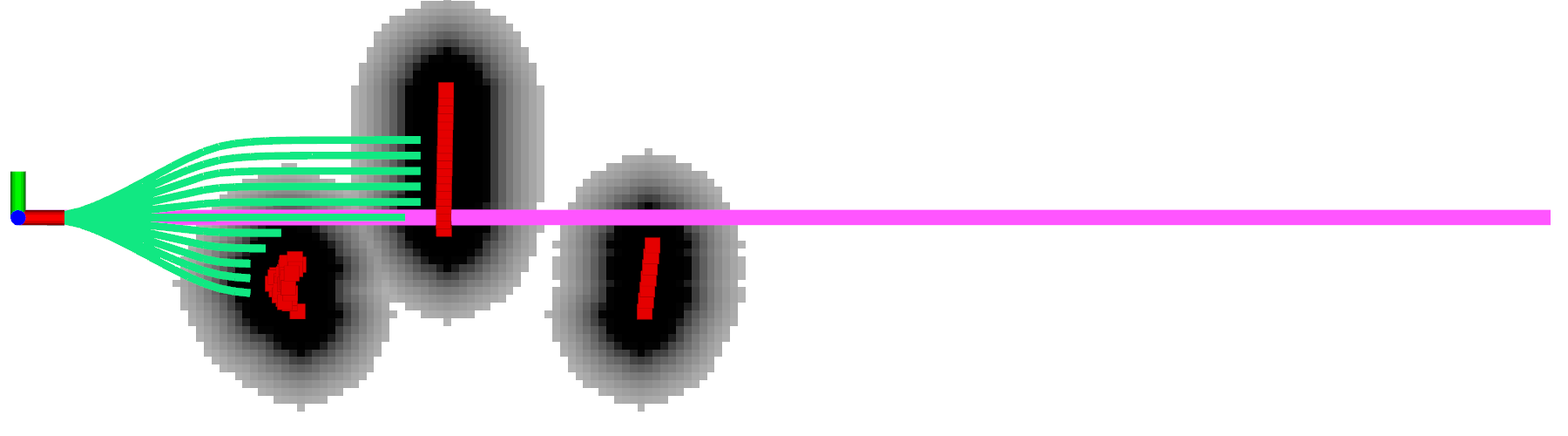}}
	\caption{Paths generated by OpenPlanner when there are (a) no obstacles, (b) one obstacle cluster, and (c) three obstacle clusters, respectively.}
	\label{fig::exp-openplanner}
	\vspace{-3mm}
\end{figure}

\begin{figure*}[ht!]
	\centering
	\subfigure[Reference path]{\includegraphics[width=.19\textwidth]{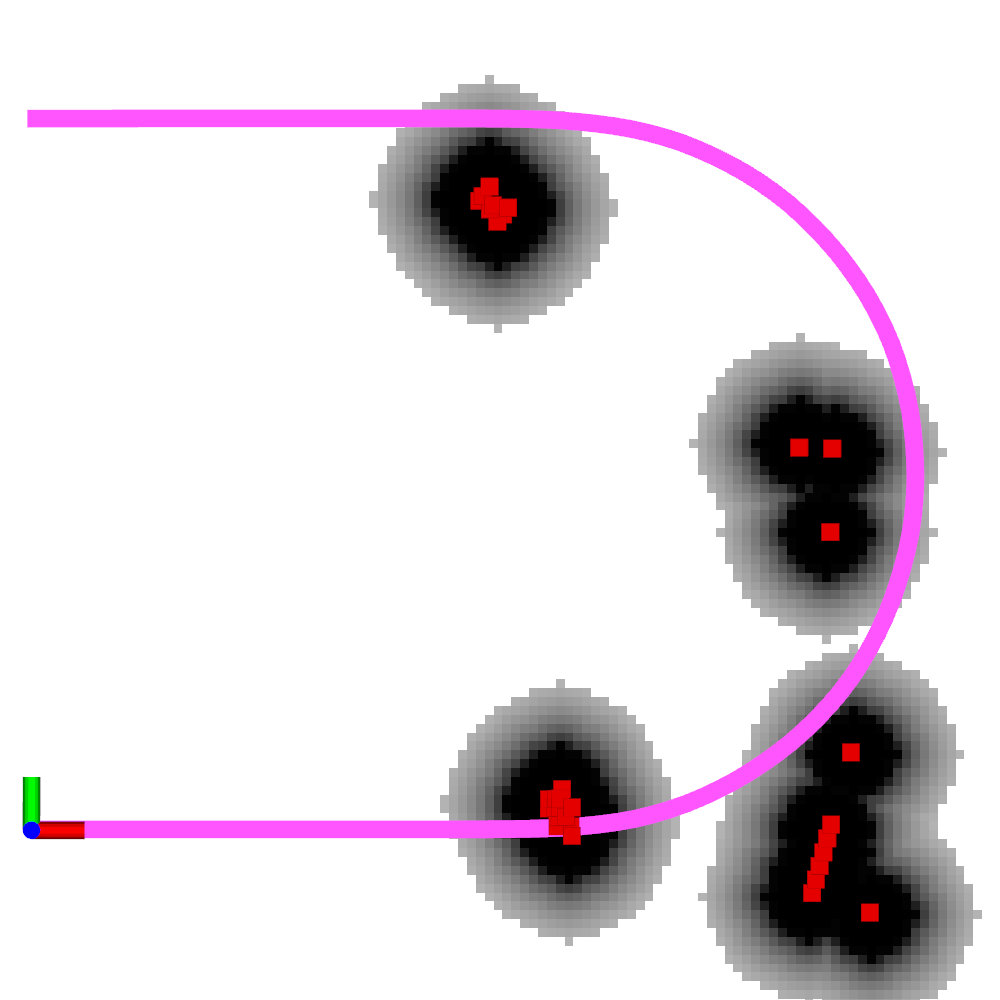}}
	\subfigure[Path of TEB planner]{\includegraphics[width=.19\textwidth]{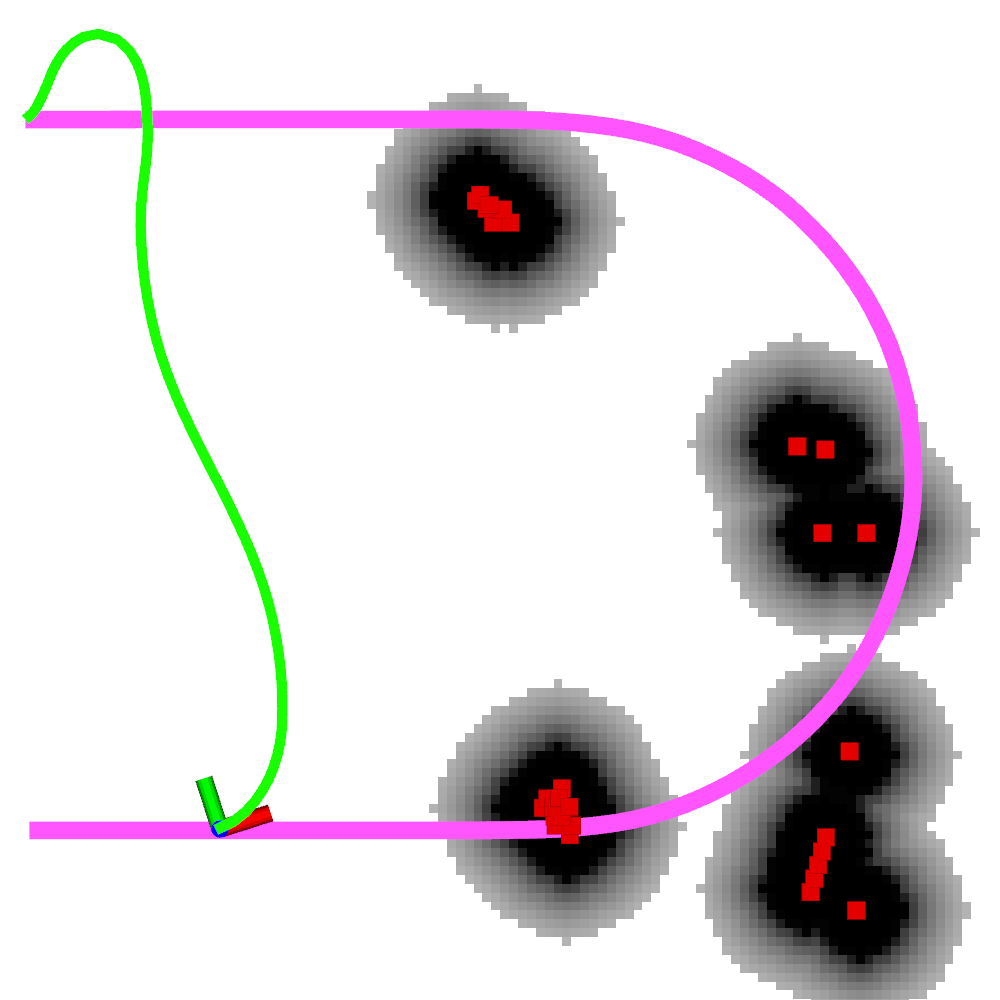}}
	\subfigure[1st path of our planner]{\includegraphics[width=.19\textwidth]{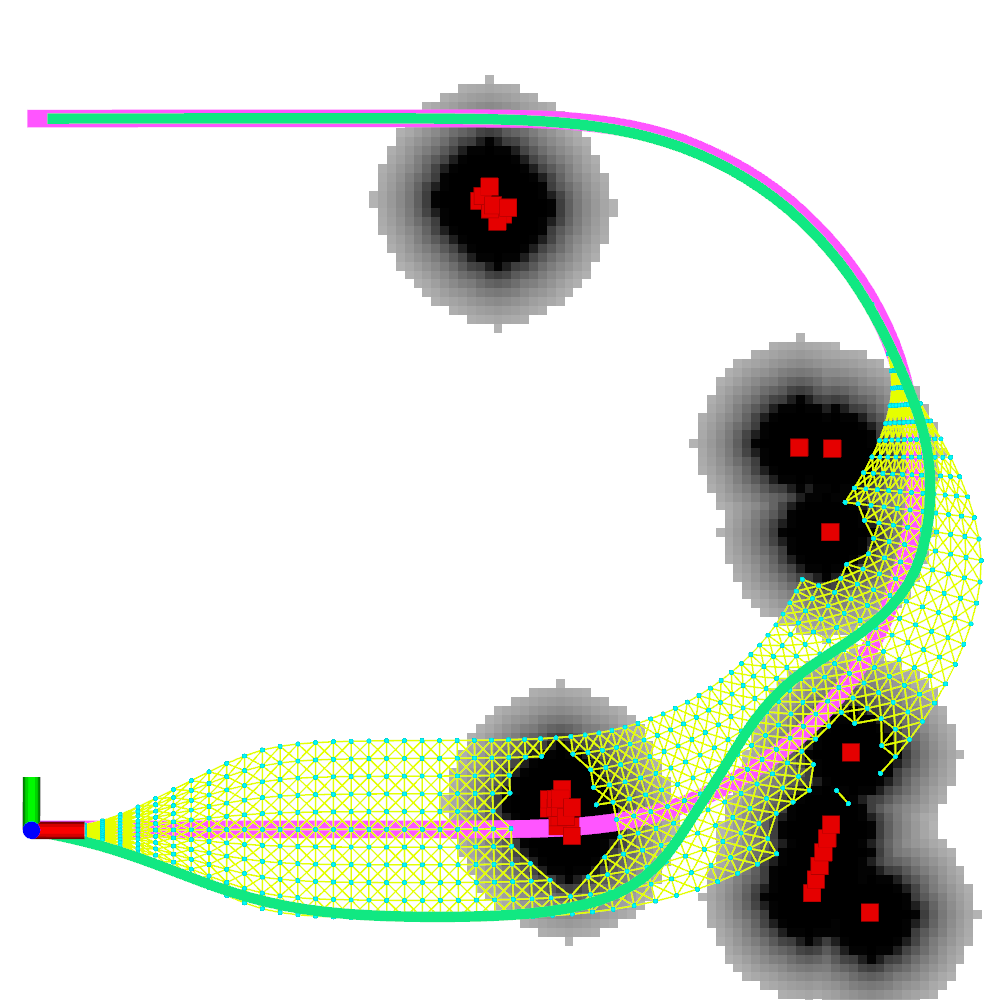}}
	\subfigure[2nd path of our planner]{\includegraphics[width=.19\textwidth]{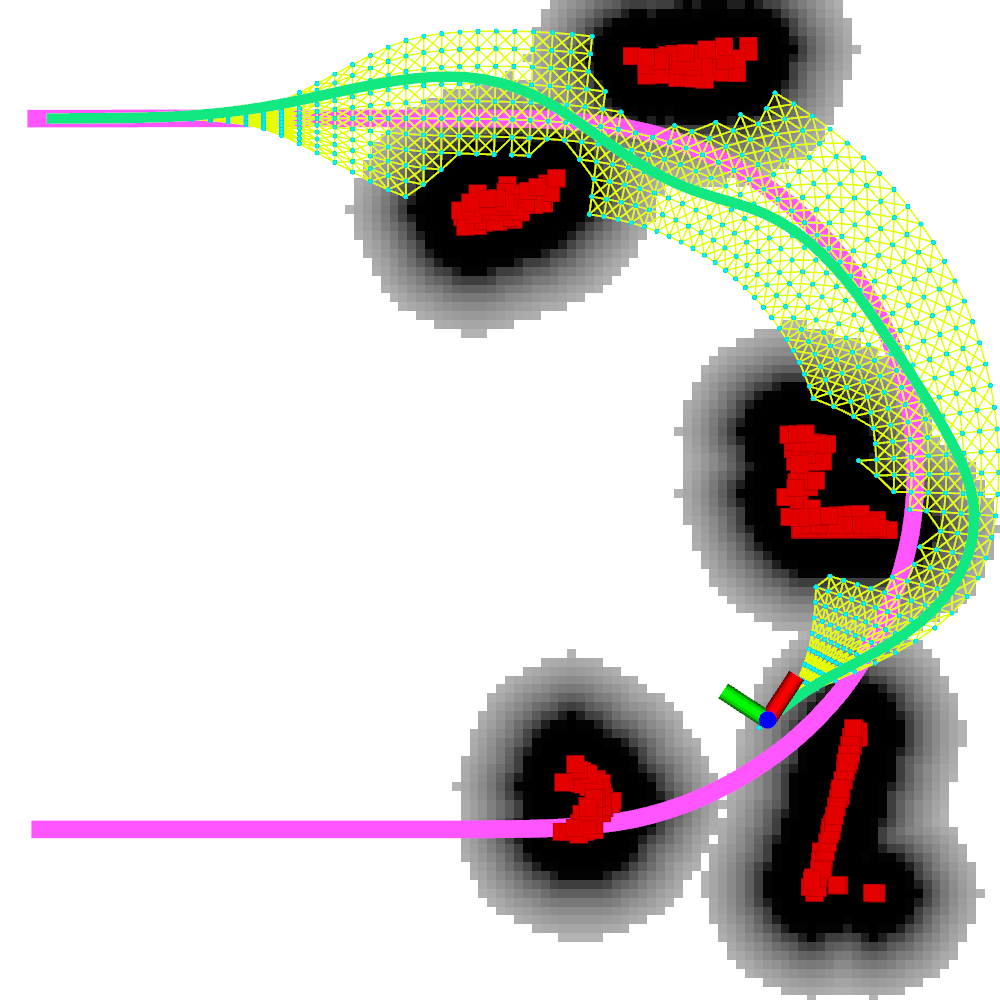}}
	\subfigure[3rd path of our planner]{\includegraphics[width=.19\textwidth]{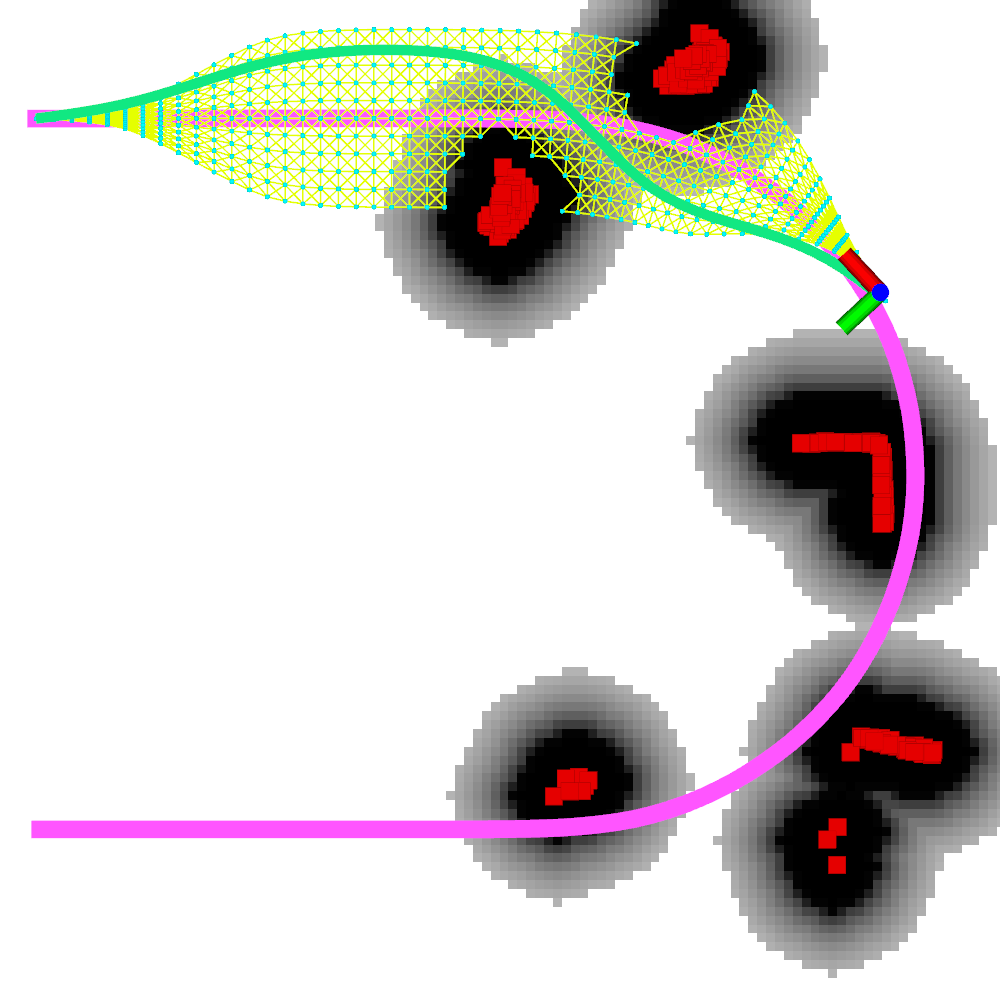}}
	\caption{Representative paths of the compared planners. In (b), the TEB planner fails to generate a path that follows the reference path. In (c)-(e), three instants are shown in which the proposed receding horizon planner avoids obstacles while adhering closely to $\mathcal{G}$.}
	\label{fig::exp-teb-planner}
	\vspace{-3mm}
\end{figure*}

\let\oldfootnotesize\footnotesize
\renewcommand*{\footnotesize}{\oldfootnotesize\scriptsize}

\section{Experiments}
\label{sec::experiment}
We now describe a series of experiments to qualitatively and quantitatively validate our proposed receding horizon planner in both simulated and real-world environments. We compare the proposed planner with OpenPlanner \cite{darweesh2017open} and Timed-Elastic-Band (TEB) Planner \cite{rosmann2012trajectory}. All compared methods are implemented in C++ and executed on a laptop equipped with an i7-10710U CPU using the robot operating system (ROS) \cite{ROS-2009} in Ubuntu Linux. We note that only the CPU is used for computation, without parallel computing enabled. Our open-source implementation of the proposed path planning framework is available on Github\footnote{\url{https://github.com/TixiaoShan/lexicographic_planning}}.

For the parameters introduced in the previous sections, we let $Th_{risk}=2$, $Th_{head}=5^{\circ}$, $d_{sensor}=5.0\;m$, $d_{roll}=7.0\;m$ for all the tests. In the simulated tests, $d_{span}$ is chosen to be 1.0 m. In the real-world experiments, $d_{span}$ is set to be 1.5 m due to the large size of the dynamic obstacles.

\subsection{Simulated Experiments}
\label{sec::experiment-sim}

\subsubsection{Comparison with OpenPlanner}
OpenPlanner \cite{darweesh2017open} is a general planning algorithm that is developed for autonomous vehicles and integrated in Autoware \cite{kato2018autoware}. Upon receiving a goal location, OpenPlanner first finds a global path using a vector map. Then local candidate roll-out paths are generated while using the global path as a reference. As is shown in Figure \ref{fig::exp-openplanner}, the roll-out paths, which are colored green, start from the vehicle location and span to cover the neighborhoods adjacent to the global reference path. The aforementioned parameter $d_{span}$ is used for defining this adjacent neighborhood. The roll-out paths are designed to be parallel to the reference path eventually. At last, a path with the lowest cost among the candidate paths will be selected and executed.

The planning scheme of OpenPlanner works well in environments with few obstacles. For a cluttered environment, it may fail to find a feasible path to execute. Such an example is shown in Figure \ref{fig::exp-openplanner}(c). All the candidates paths of OpenPlanner are blocked by three clusters of obstacles on the global reference path. On the other hand, the proposed receding horizon planner does not  encounter such a problem because we construct a more comprehensive graph and search for feasible paths. The returned path of our planner over this example is shown in Figure \ref{fig::pipeline}(c).

\subsubsection{Comparison with TEB}
TEB \cite{rosmann2012trajectory} is an optimization-based planner that also takes a global path as a reference. It generates an executable path by deforming the reference path while taking the dynamic constraints of the robot into account. The optimization problem of the TEB planner is formulated as a weighted-sum multi-objective problem, where the weights are manually adjusted by the user.

In this test, we compare the proposed receding horizon planner with the TEB planner by following a U-shaped global path, which is shown in Figure \ref{fig::exp-teb-planner}(a). The environment is populated with randomly placed obstacles around the reference path. The path returned by the TEB planner is shown in Figure \ref{fig::exp-teb-planner}(b). Its path skips the entire operational region and leads the robot directly to the goal, which goes against the original intention of following a reference path.

When we test the proposed planner, the first path returned is shown in Figure \ref{fig::exp-teb-planner}(c). When the robot moves forward by following this path, new obstacles (shown in the mid-right and mid-top of Figure \ref{fig::exp-teb-planner}(d)) are detected on the path being executed. Thus the first path becomes invalid. Re-planning is performed and yields the second path (Figure \ref{fig::exp-teb-planner}(d)). As the robot explores the environment more, a safer path with lower risk cost is found and returned as the third path shown in Figure \ref{fig::exp-teb-planner}(e). During the entire run, our planner only re-plans three times and traverses safely amid the obstacles, while staying close to the reference path.

\begin{figure}[ht]
	\centering
	\includegraphics[width=.99\columnwidth]{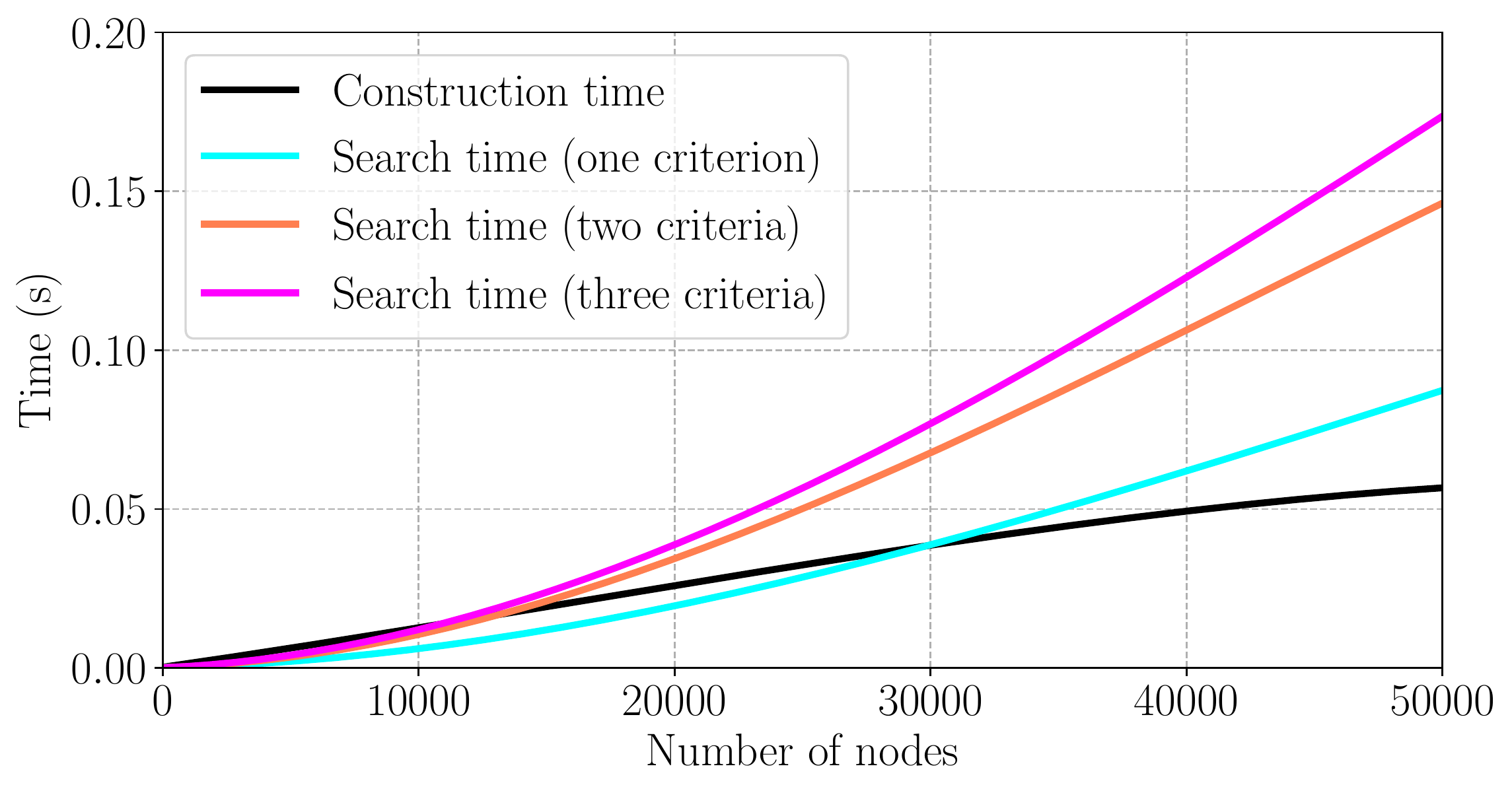}
	\caption{Algorithm runtime as a function of the number of nodes in the graph when performing graph construction and search. The construction time is primarily devoted to sampling kinodynamic states and edge collision checking using the sensor data.}
	\label{fig::benchmarking-time}
	\vspace{-3mm}
\end{figure}

\subsubsection{Benchmarking}
We show the algorithm runtime of the proposed planner in Figure \ref{fig::benchmarking-time}. The graph construction time, which scales linearly as the number of the nodes increases, is plotted in black. In order to explore the influence of introducing new costs into the lexicographic ordering on the planning performance, we show benchmarking results using three cost combinations: (a) one criterion - distance only, (b) two criteria - a heading-distance ordering, and (c) three criteria - a risk-heading-distance ordering. The lexicographic search times when using these three cost combinations are depicted in cyan, orange, and pink respectively in Figure \ref{fig::benchmarking-time}.

\begin{figure}[h]
	\centering
	\subfigure[]{\includegraphics[width=.32\columnwidth]{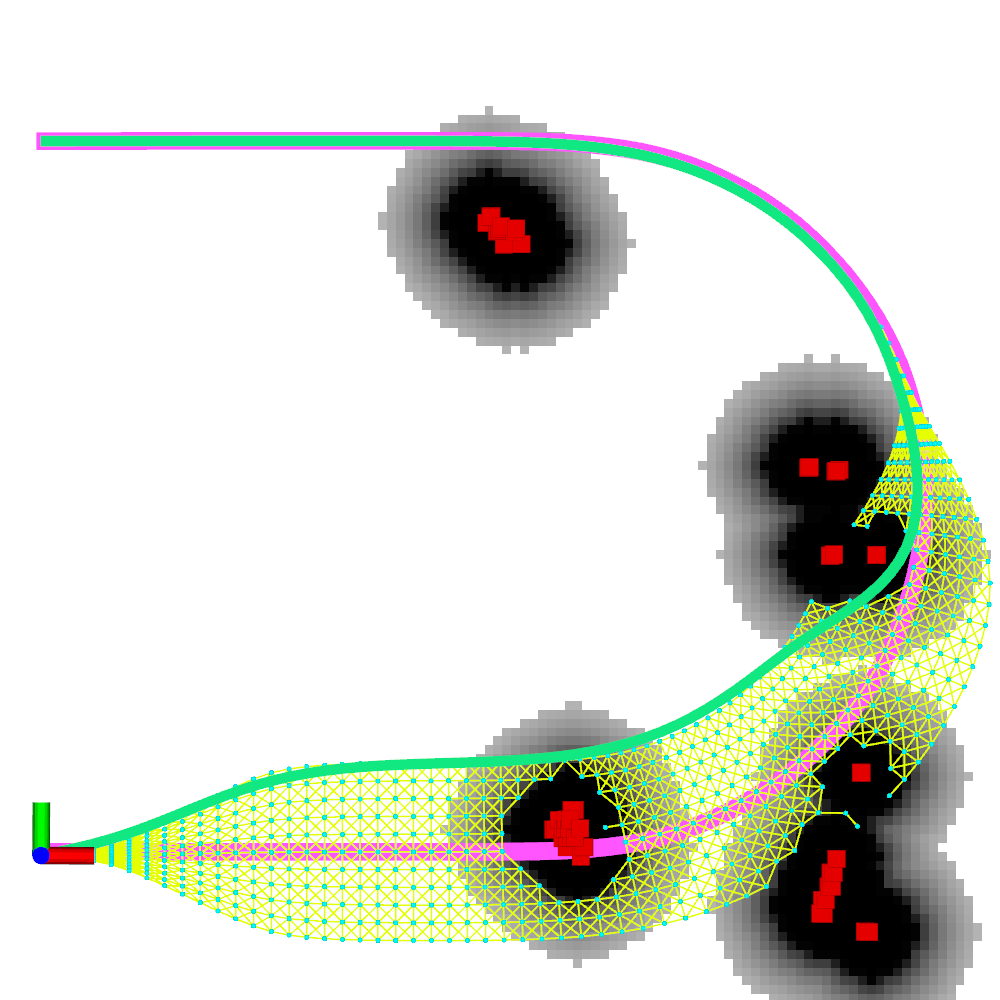}}
	\subfigure[]{\includegraphics[width=.32\columnwidth]{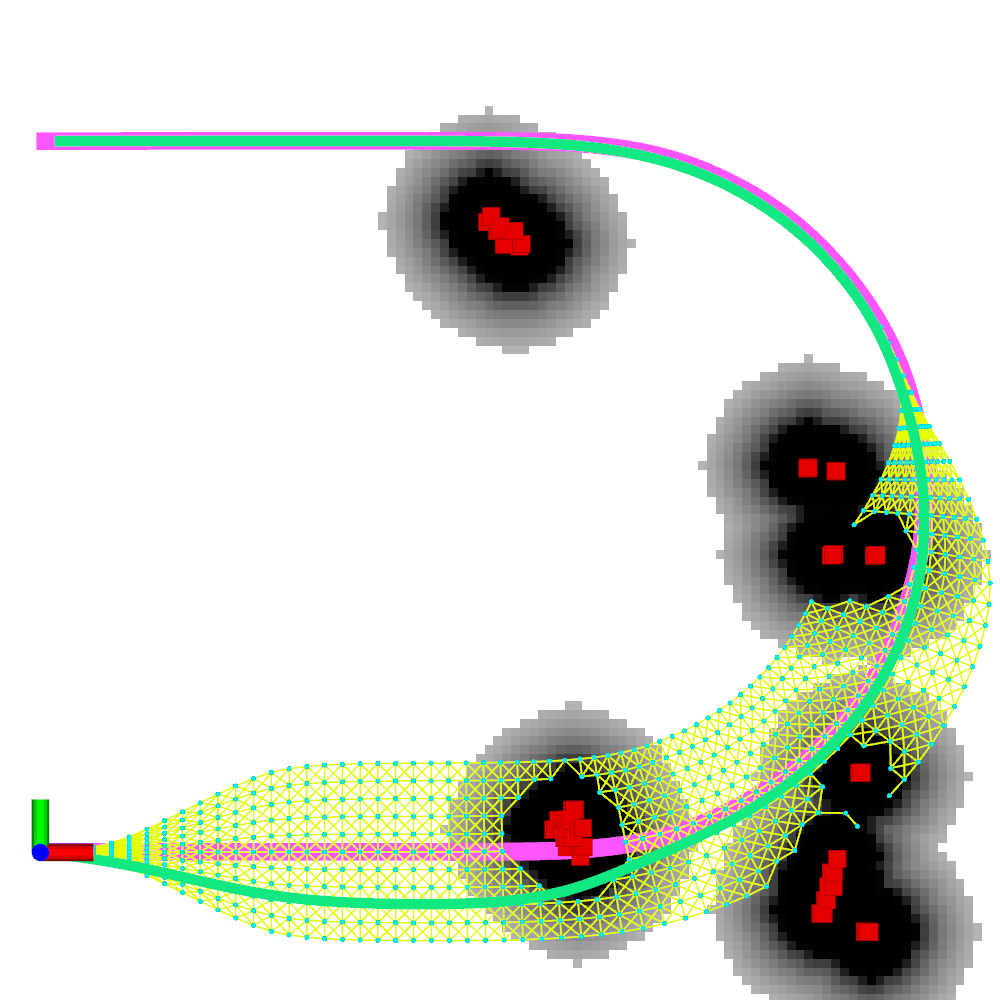}}
	\subfigure[]{\includegraphics[width=.32\columnwidth]{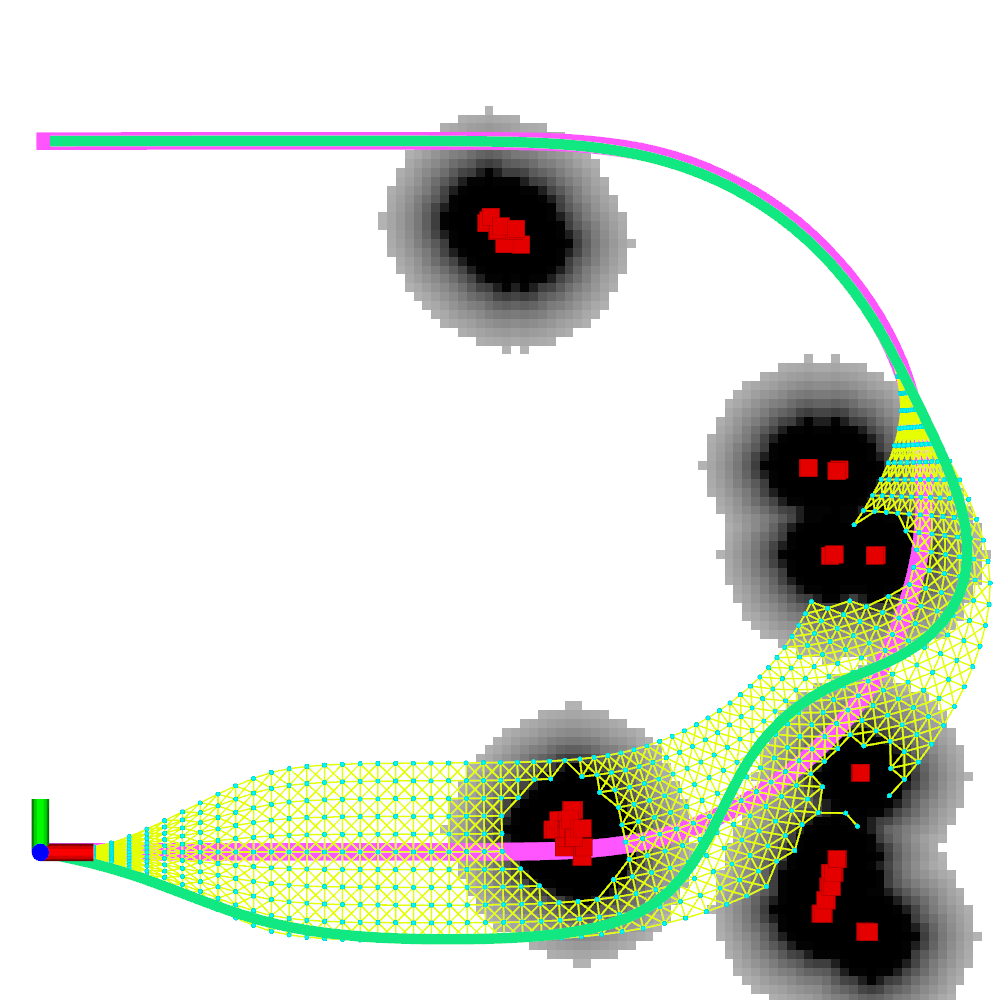}}
	\vspace{-3mm}
	\caption{Paths generated using different criteria combinations: (a) only distance cost is applied during graph search, (b) heading and distance costs are used, (c) risk, heading, distance costs are used.}
	\label{fig::benchmarking-path}
	\vspace{-3mm}
\end{figure}

Figure \ref{fig::benchmarking-path} shows the corresponding solutions when applying three different cost combinations using the test environment illustrated in Figure \ref{fig::exp-teb-planner}(a). Table \ref{tab::benchmarking-cost} shows the values of the corresponding costs of these three combinations. Note that the cost values marked in parentheses in Table \ref{tab::benchmarking-cost} are calculated for reference and not used in the optimization process. When only distance cost is applied, the shortest path is found and shown in Figure \ref{fig::benchmarking-path}(a). This path starts with swinging to the left and ends with converging back to the global path. Though this path achieves the lowest distance cost of the three cost combinations, it yields very high risk cost, as it stays close to the obstacles. Figure \ref{fig::benchmarking-path}(b) shows the path when we apply two cost criteria. After adding the heading cost to the hierarchy, the obtained path possesses fewer heading differences from the reference trajectory even as the distance cost increases. As is shown in Figure \ref{fig::benchmarking-path}(c), the utilization of a risk-heading-distance hierarchy yields the safest path by keeping away from surrounding obstacles. Due to space limitations, we refer the reader to \cite{shan2019lexicographic} for a more detailed computational analysis of lexicographic search under different quantities and combinations of costs, and varying node densities in the graph.

\begin{table}[h!]
	\caption{Costs when applying different criteria combinations}
	\label{tab::benchmarking-cost}
	\centering
	\resizebox{0.9\columnwidth}{!}{
	\begin{tabular}{cccc}
		\toprule
		Cost     & One criterion & Two criteria & Three criteria \\
		\midrule
		Risk     & (86.8)               & (78.3)                & 32.7        \\
		Heading  & (35.4$^{\circ}$)     & 19.0$^{\circ}$        & 56.4$^{\circ}$           \\
		Distance & 6.8 m                 & 7.4 m                  & 8.1 m         \\
		\bottomrule
	\end{tabular}
	}
	\vspace{-3mm}
\end{table}

\subsection{Real-world Experiments}
\label{sec::experiment-real}

\begin{figure}[t!]
	\centering
	\subfigure[]{\includegraphics[width=.6\columnwidth]{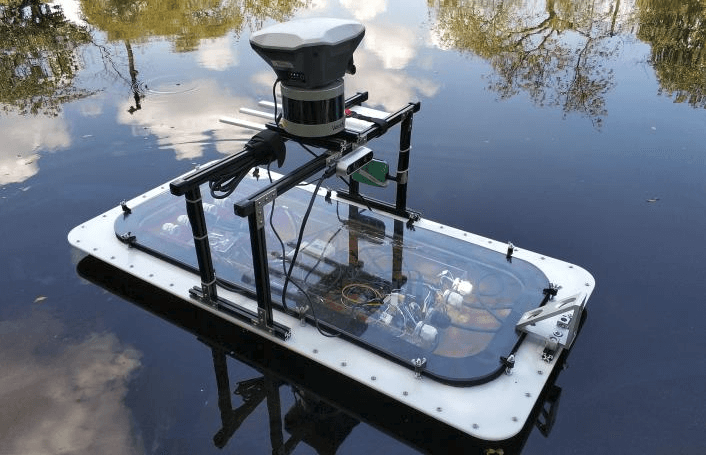}}
	\subfigure[]{\includegraphics[width=.2\columnwidth]{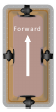}}
	\caption{Quarter-scale Roboat hardware and thruster placement.}
	\label{fig::roboat-hardware}
	\vspace{-3mm}
\end{figure}

\begin{figure}[ht!]
	\centering
	\subfigure[]{\includegraphics[width=.32\columnwidth]{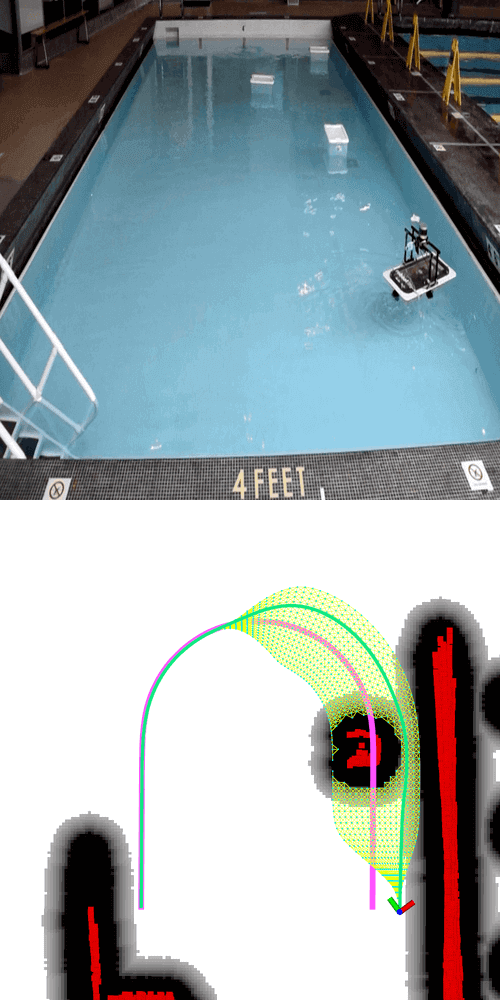}}
	\subfigure[]{\includegraphics[width=.32\columnwidth]{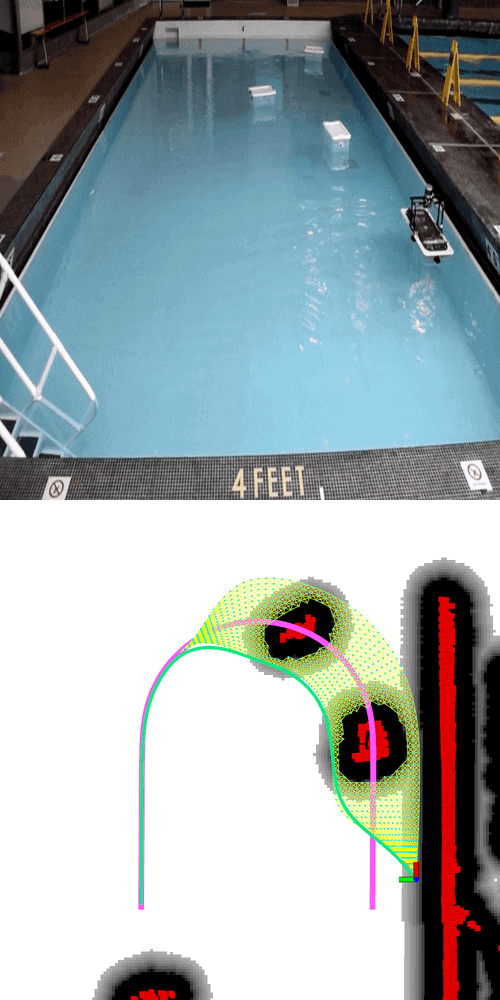}}
	\subfigure[]{\includegraphics[width=.32\columnwidth]{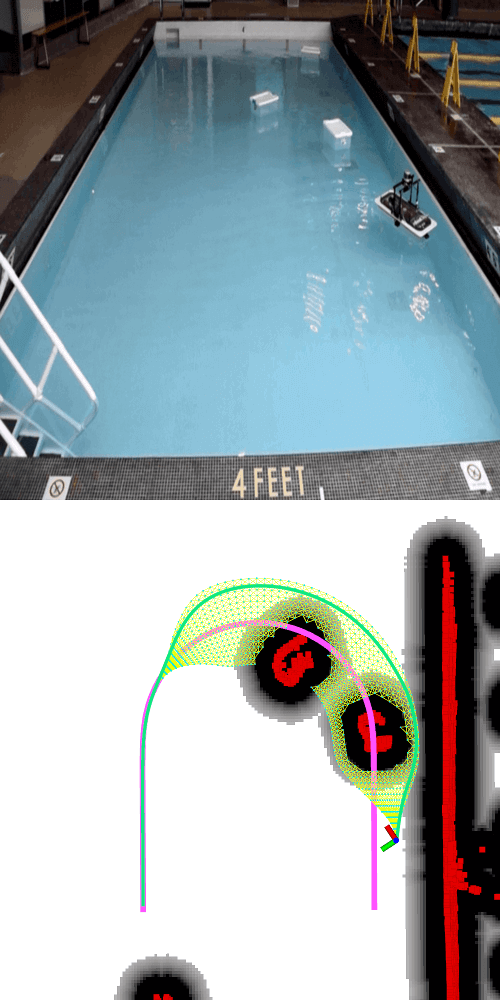}}
	\subfigure[]{\includegraphics[width=.32\columnwidth]{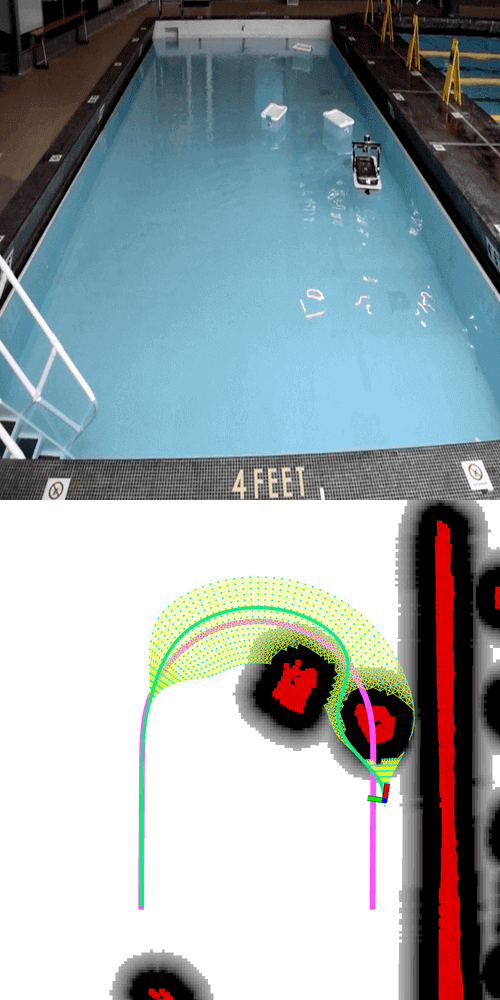}}
	\subfigure[]{\includegraphics[width=.32\columnwidth]{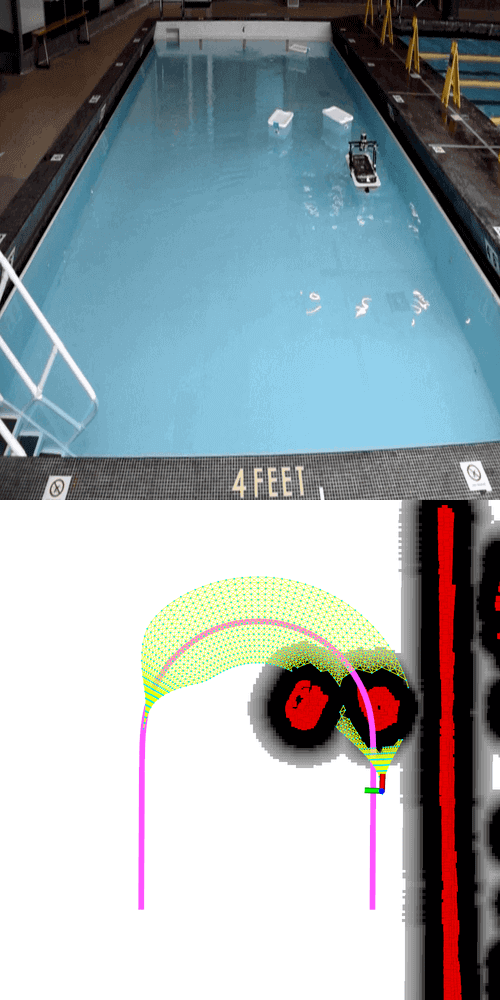}}
	\subfigure[]{\includegraphics[width=.32\columnwidth]{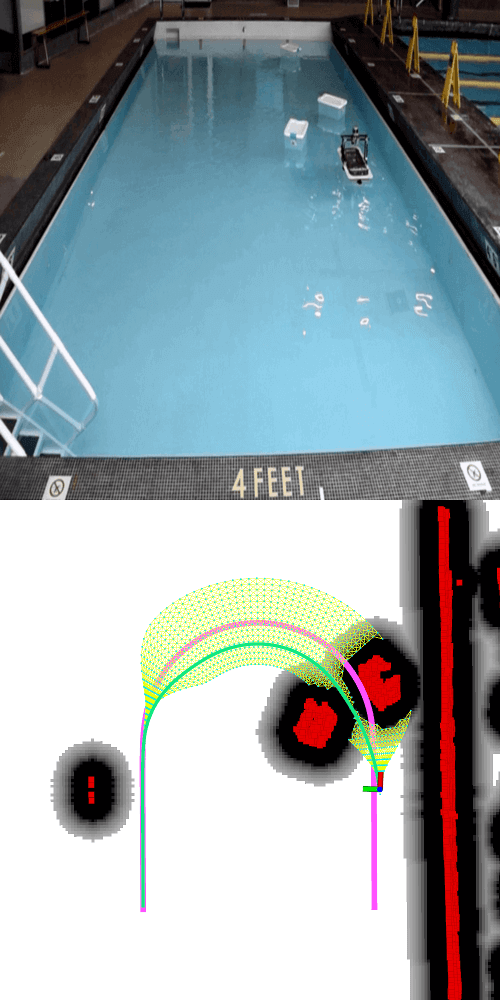}}
	\caption{Testing the proposed receding horizon planner on a quarter-scale Roboat in a swimming pool, with moving obstacles. The robot starts from the lower-right corner of the figure and tries to reach the goal at the lower-left corner. The reference path is colored pink.}
	\label{fig::exp-pool-2}
	\vspace{-3mm}
\end{figure}

Finally, we implement the proposed receding horizon planner on an autonomous surface vehicle - the quarter-scale Roboat \cite{wang2019roboat}. As is shown in Figure \ref{fig::roboat-hardware}, Roboat has dimensions of 0.9 m $\times$ 0.45 m $\times$ 0.5 m (L$\times$W$\times$H) and weighs about 15 kg. It is outfitted with four thrusters as shown in Figure \ref{fig::roboat-hardware} to enable omnidirectional maneuvering, and it is equipped with a Velodyne VLP-16 for perception. Localization is performed using a modified lidar odometry framework adapted from \cite{shan2020lio}. We discard laser range returns that are more than 5 meters away from the robot, allowing a small-scale environment to produce varied outcomes. The laser range returns within 5 meters are considered to be readings from obstacles.

We conduct an experiment in a 12.5 m $\times$ 6.5 m swimming pool. Three floating containers are placed in the pool to serve as obstacles. The containers move randomly in the pool due to water flow. The U-shaped global reference path is colored pink in Figure \ref{fig::exp-pool-2}. Besides the containers, other structures, such as walls, in the environment are also considered as obstacles. The robot starts from the lower-right corner of the figure and tries to reach the goal at the lower-left corner. At the beginning of planning (Figure \ref{fig::exp-pool-2}(a)), only one container is within the sensor range and on the reference path. The returned path swings to the right to avoid it. In Figure \ref{fig::exp-pool-2}(b), the path changes accordingly when another container is detected by the robot. Due to the moving obstacles, the returned path from our planner changes several times (Figure \ref{fig::exp-pool-2} (b)-(d)). Note that these paths strictly follow the cost-hierarchy we defined in Section \ref{sec::problem-cost}. As is shown in Figure \ref{fig::exp-pool-2}(e), two obstacles completely block the forward path of the vehicle. The robot remains stationary and waits for the waterway to clear. In Figure \ref{fig::exp-pool-2}(f), a new path is found as one of the obstacles moves away. The robot follows this path and reaches the goal location eventually.

\section{Conclusions and Discussion}
We have proposed a receding horizon planner for path planning with an ASV in urban waterways. The receding horizon planner generates a graph from a global reference path to search for feasible paths in the presence of obstacles. We also propose a lexicographic search method intended for use with graphs in multi-objective robot motion planning problems, in which competing resources are penalized hierarchically. Over such problems, we have demonstrated that the proposed search method is capable of producing high-quality solutions with efficient runtime. The variant of Dijkstra's algorithm proposed for performing the search offers appealing scalability, as its worst-case complexity scales linearly in the number $K$ of cost criteria. A key benefit of the approach is that, in contrast to planning methods that employ weight coefficients or constraints, minimal tuning is required, beyond the ordering of cost functions in the hierarchy. Since no constraints other than obstacle avoidance need be imposed, feasible solutions are obtained quickly. Real-world implementation of our method is also demonstrated on an ASV. Future work entails the extension of this method to time-varying costs that are history-dependent, for use in motion planning under uncertainty.

\section*{Acknowledgement}

This work was supported by Amsterdam Institute for Advanced Metropolitan Solutions, Amsterdam, the Netherlands.

\bibliographystyle{IEEEtran}
\bibliography{CDC_2020_Tixiao}
\end{document}